# Egocentric Spatial Memory

Mengmi Zhang, Keng Teck Ma, Shih-Cheng Yen, Joo Hwee Lim, Qi Zhao, and Jiashi Feng

*Abstract*— Egocentric spatial memory (ESM) defines a memory system with encoding, storing, recognizing and recalling the spatial information about the environment from an egocentric perspective. We introduce an integrated deep neural network architecture for modeling ESM. It learns to estimate the occupancy state of the world and progressively construct top-down 2D global maps from egocentric views in a spatially extended environment. During the exploration, our proposed ESM model updates belief of the global map based on local observations using a recurrent neural network. It also augments the local mapping with a novel external memory to encode and store latent representations of the visited places over long-term exploration in large environments which enables agents to perform place recognition and hence, loop closure. Our proposed ESM network contributes in the following aspects: (1) without feature engineering, our model predicts free space based on egocentric views efficiently in an end-to-end manner; (2) different from other deep learning-based mapping system, ESMN deals with continuous actions and states which is vitally important for robotic control in real applications. In the experiments, we demonstrate its accurate and robust global mapping capacities in 3D virtual mazes and realistic indoor environments by comparing with several competitive baselines.

## I. INTRODUCTION

Egocentric spatial memory (ESM) refers to a memory system that encodes, stores, recognizes and recalls the spatial information about the environment from an egocentric perspective [20]. Such information is vitally important for embodied agents to construct spatial maps and reach goal locations in navigation tasks.

In this paper, we formulate egocentric spatial memory as a mapping task where the agent with ESM takes pre-planned egomotions in a continuous action space and constructs an accurate top-down global occupancy grid map based on egocentric camera views. In robotics, the idea of occupancy grid map originated from [5] to model the occupancy level of a static environment by dividing the world into a fixed sized grid and then updating the occupancy probability of each cell individually. Since then, a reliable occupancy map has been an imperative aspect in path planning and navigation in unmapped indoor and outdoor environments, such as autonomous driving cars [14].

Recently, a wealth of neurophysiological results have shed lights on the underlying neural mechanisms of ESM in mammalian brains mostly through single-cell electrophysiological recordings in mammals [21], [32], [4], [2], [30], [12]. There are four types of cells identified as specialized for processing spatial information: head-direction cells, boundary vector cells, place cells and grid cells. Inspired by neurophysiological discoveries on these four types of navigation cells, we propose the computational architecture, named as the Egocentric Spatial Memory Network (ESMN), for modeling ESM. ESMN encapsulates the four cell types respectively with functionally similar neural network-based modules named as Head Direction Unit, Boundary Vector Unit, Place Unit and Grid Unit. However, we emphasize that these are descriptive naming conventions rather than firm biological equivalents.

In the architecture design, ESMN fuses the egocentric observations from the agent over time and produces a top-down 2D local map using a recurrent neural network. To align the spatial information at the current step with all the past predicted local maps, ESMN transforms all the past information using a spatial transformer neural network based on the agent's pre-planned egomotion. ESMN also augments the local mapping module with a novel external memory capable of integrating local maps into global maps and storing the discriminative representations of the visited places. Although we introduce noise in pre-planned egomotions to demonstrate the usefulness of external memory of our model in loop closure classification and map correction in the experiment section; it is important to make it clear that, *different* from Simultaneous Localization and Mapping (SLAM), our focus is to propose an end-to-end deep network-based mapping framework given the ground truth egomotions in the continuous action space, *i.e.*, the agent has full knowledge about its current state based on pre-planned egomotions.

We first evaluate ESMN in 3D maze environments where they feature complex geometry and varieties of textures. After fine tuning ESMN, we assess its long-term mapping efficacy over large areas in realistic indoor environments from 2D3DS dataset [1]. Experimental results demonstrate the acquired skills of ESMN in terms of free space prediction and long-term map construction over large areas in indoor environments. Lastly, we perturb the pre-planned egomotions with noise and show the usefulness of our external memory in loop closure classification using precision versus recall and map correction by comparing with other baselines.

## II. RELATED WORKS

There is a rich literature on computational models of egocentric spatial memory (ESM) primarily in cognitive

Mengmi Zhang is with National University of Singapore (NUS), Singapore. Email: mengmi@u.nus.edu

Keng Teck Ma and Joo Hwee Lim are with A*AI, SCEI, I2R, A*STAR, Singapore. Email: ma_ken_teck@scei.a-star.edu.sg, joohwee@i2r.a-star.edu.sg

Qi Zhao is with University of Minnesota, USA. Email: qzhao@umn.edu

Shih-Cheng Yen and Jiashi Feng are with NUS, Singapore. Email: {shihcheng, elefjia}@nus.edu.sg

science and AI. For brevity, we focus on the related works in machine learning and robotics.

Reward-based learning is frequently observed in spatial memory experiments [15]. In machine learning, reinforcement learning is commonly used in attempts to mathematically formalize the reward-based learning. Deep Q-networks, one of its frameworks, have been employed in the navigation tasks where the agent aims to maximize the rewards while navigating to the goal locations [23], [26], [25]. The representation of spatial memory is expressed implicitly using either long short term memory (LSTM)[13] or addressable memory[28], [34]. In one of the relevant works [9] and its following work [10], Gupta *et al.* introduced a mapper-planner pipeline where the agent is trained in a supervised way to produce a top-down belief map of the world and thus plans path; however, the authors assume that the agents perform mapping tasks with limited area coverage in an ideal scenario where the agent could pick one out of the six discrete actions every time step. However, robotics are often connected with continuous actions and states in real applications. Different from their work, our ESM network extends to continuous action space. Moreover, we augment the local mapper as described in [9] with a novel external memory for storing discriminative representation of visited places which helps us correct integration errors over long terms in large areas.

Besides reinforcement learning, there are also works on navigation in robotics where the spatial memory is often explicitly represented in the form of grid maps or topological structures[5], [29], [19], [22]. As great strides have been made using deep learning in computer vision tasks which results in a significant performance boost, several works [31], [11], [27] endeavor to use deep learning to construct occupancy maps from range sensors, such as LIDAR. However, there is no existing end-to-end deep neural network for global occupancy grid map construction based on egocentric camera views to our best knowledge.

## III. PROPOSED ARCHITECTURE

We first introduce ESM modeling problem and relevant notations. ESM involves an object-to-self representational system which constantly requires encoding, transforming and integrating spatial information from the first-person view into a global map. At the current time step $t$, the agent, equipped with one RGB camera, takes the camera view $I_t$ and the pre-planned egomotion measured from motion sensors as the current inputs. We assume the measurements from motion sensors are noise-free, *i.e.*, the agent's next state can be accurately derived from integration of egomotion with the previous state. As place recognition and memory recall are critical aspects of ESM, in the experiment section, we demonstrate examples where we introduce noise to the motion measurements and how we use external memory storing discriminative place embedding for loop closure detection and hence, map correction.

At time step $t$, the agent is allowed to take one egomotion out of a continuous set of actions which include rotating left/right by $\theta$ degrees and moving in the directions $l$ by the distance of $d$ relative to the agent's current pose. We assume the agent moves freely in a 2D world and the camera coordinate is fixed with respect to the agent's body. The starting location $p_0$ of the agent is always $(0,0,0)$ where the triplet denotes the positions along the x and y-axis and the orientation in the world coordinate. The problem of modeling ESM is to learn a global map in the egocentric coordinate based on the visual input $I_t$ for $t = 1, \ldots, T$ and motion sensor measurements. We define the global map as a top-view 2D probabilistic grid map where the probability infers the agent's belief of free space. In order to tackle this problem, we propose a unified neural network architecture named as ESMN.

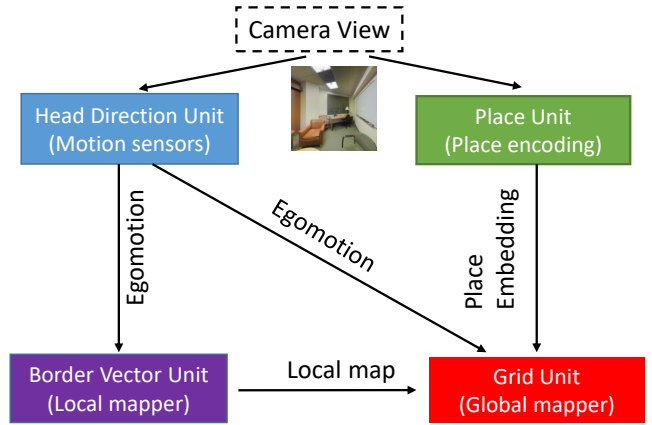

Fig. 1: Overview of our proposed Egocentric Spatial Memory Network. It consists of Head Direction Unit, Boundary Vector Unit, Place Unit, and Grid Unit. See Section III-A for more details.

### A. Overview

The architecture of ESMN is illustrated in Figure 1 and is elaborated in details in Figure 2. Inspired by the navigation cells mentioned in the introduction, our proposed ESMN comprises four modules: Head Direction Unit (HDU), Boundary Vector Unit (BVU), Place Unit (PU), and Grid Unit (GU). This is to incorporate multiple objectives for modeling ESM. (1) We simplify HDU as the motion sensors and it outputs the egomotion measurement at time step $t$. (2) BVU serves as a local mapper and predicts 2D top-view local maps representing free space. It minimizes errors of predicted free space in the local map by using a recurrent neural network. Based on the egomotion, a spatial transformer module transforms all the past spatial information to the current egocentric coordinate via a 2D-CNN. (3) PU learns to encode the latent representation of visited places in a 2D-CNN pre-trained using a triplet loss. (4) GU, as external memory, integrates the predicted local maps from BVU over time and keeps track of all the visited places.

In this paper, we adopt a "divide and conquer" approach by composing different navigation modules within one framework systematically. Leveraging on rich features extracted from deep networks, the learnt features are suitable for recognizing visual scenes and hence, boost up the map construction and correction performances. The efficiency and robustness of our algorithm are demonstrated by comparing with other spatial mappers based on visual inputs.

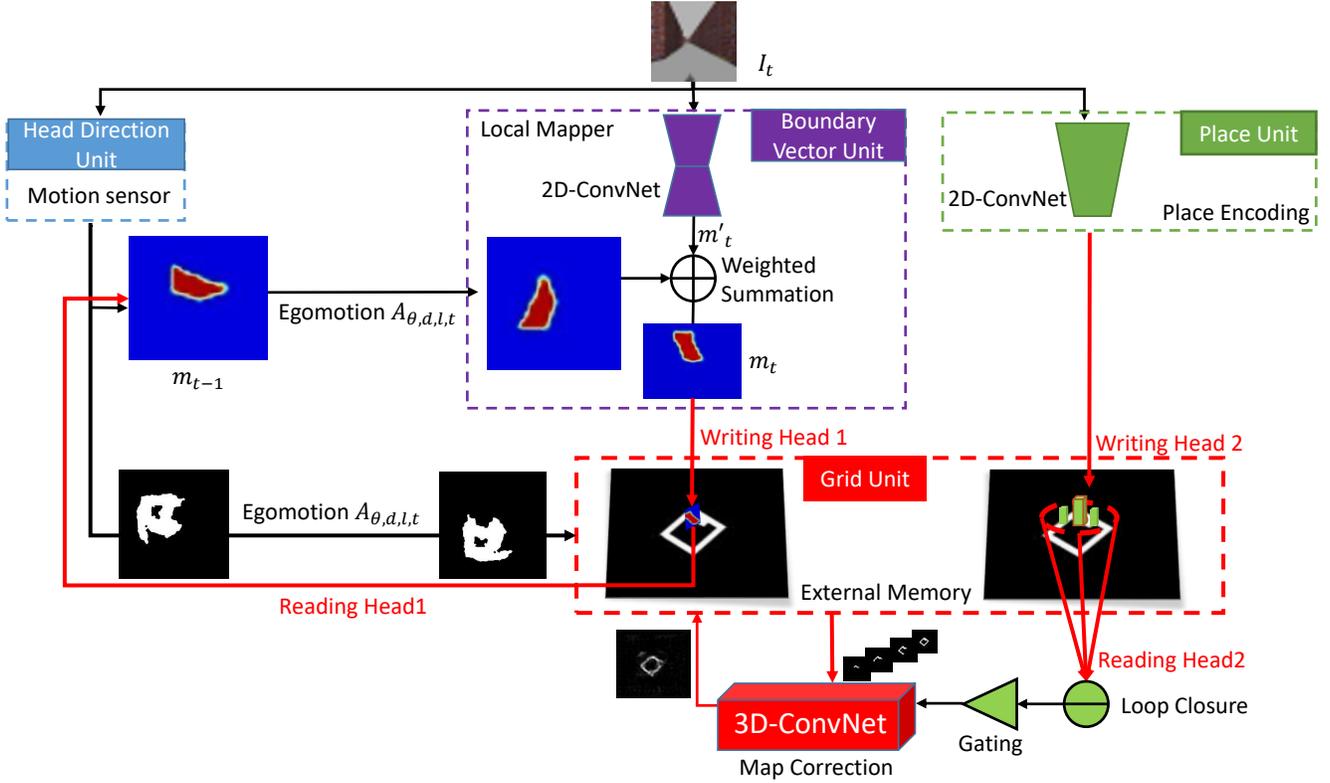

Fig. 2: Architecture of our Egocentric Spatial Memory Network. See Section III-A for the overview of individual module.

### B. Head Direction Unit (HDU): egomotion measurement

At time step $t$, the current RGB camera image (or frame) $I_t$ is the external input to the agent. In order to make spatial reasoning about the topological structure of the spatially extended environment, the agent has to plan paths to explore its surroundings and predict their own poses by integrating the egomotions over time. As path planning and pose estimation themselves can be research topics, we simplify the HDU module in ESMN with an ideal motion sensor which integrates its measurement over time and returns noise-free egomotion $A_{\theta,d,l,t}$. In other words, the artificial agent has full knowledge about the localization information on the pre-planned trajectory which is obtained via integration of ground-truth egomotions. See Section IV-C in the experiment section for an example where we introduce measurement noise from motion sensors and demonstrate how individual module contributes to robust mapping.

### C. Boundary Vector Unit (BVU): local mapper

We explain how ESMN integrates egocentric views into a top-down 2D representation of the environment using a recurrent neural network. Similar to [9], the BVU in ESMN serves as a local mapper and maintains the accumulative free space representations in the egocentric coordinate for a short-term period with fixed map sizes. Given the current observation $I_t$, function $g$ first encodes its geometric representation about space $g(I_t)$ via a 2D-CNN and then transform $g(I_t)$ into egocentric top-down view $m'_t$ via deconvolutional layers. Together with the accumulative space representation $m_{t-1}$ at the previous step and the estimated egomotion $A_{\theta,d,l,t}$ from $t-1$ to $t$, BVU estimates the current local space representation $m_t$ using the following rule:

$$m_t = U(S(m_{t-1}, A_{\theta,d,l,t}), m'_t), \quad (1)$$

where $S$ is a function that transforms the previous accumulative space representation $m_{t-1}$ to the current egocentric coordinate based on the measured egomotion $A_{\theta,d,l,t}$. We parameterize $S$ by using a spatial transformer network [16] composing of two key elements: (1) it generates the sampling grid which maps the input coordinates of $m_{t-1}$ to the corresponding output locations after egomotion $A_{\theta,d,l,t}$ transformation; (2) the sampling kernel then takes the bilinear interpolation of the values on $m_{t-1}$ and outputs the transformed $m_{t-1}$ in the current egocentric coordinate. $U$ is a function which merges the free space prediction $m'_t$ from the current observation with the accumulative free space representation at the previous step. Specifically, we simplify merging function $U$ as a weighted summation parameterized by $\lambda$ followed by hyperbolic tangent operation:

$$U(S(m_{t-1}, A_{\theta,d,l,t}), m'_t) = \frac{e^{2(\lambda m'_t + (1-\lambda)S(m_{t-1}, A_{\theta,d,l,t}))} - 1}{e^{2(\lambda m'_t + (1-\lambda)S(m_{t-1}, A_{\theta,d,l,t}))} + 1}, \quad (2)$$

## D. Place Unit (PU): place encoding

One of the critical aspects of ESM is the ability to recall and recognize a previously visited place independent of agent's orientation. To eliminate the accumulated errors during long-term mapping, loop closure is valuable for the agents in the navigation tasks. In order to detect loop closure during an episode, given the current observation $I_t$, ESMN requires to encode the discriminative representation $h(I_t)$ of specific places independent of scaling and orientations via an embedding function $F$. Based on the similarity of all the past observations $\Omega_t = \{I_1, I_2, ..., I_t\}$ at corresponding locations $P_t = \{p_1, p_2, ..., p_t\}$, we create training targets by making an analogy to the image retrieval tasks [7] and define the triplet $(I_t, I_+, I_-)$ as anchor sample (current camera view), positive and negative samples drawn from $\Omega_t$ respectively. PU tries to minimize the triplet loss:

$$L_{triplet}(F(I_t, I_+, I_-)) = \\ -\log \frac{e^{-D(F(I_t, I_+))}}{e^{-D(F(I_t, I_+))} + e^{-D(F(I_t, I_-))}}, \quad (3)$$

where we parameterize $F$ using a three-stream 2D-CNN where the weights are shared across streams. $D$ is a distance measure between pairs of embedding. Here, mean squared error is used. See Section IV-B and Section IV-C for examples where we perturb the pre-defined trajectory for the agent with noise. Given the current camera view, PU identifies the previous camera view where the loop closure was detected based on the encoded place representations.

## E. Grid Unit (GU): global mapper and place tracking

While BVU provides accumulative local free space representations in high resolution for a short-term period, we augment the local mapping framework with external memory for long-term integration of the local maps and storage of location representations. Compared with the local map $m_t$ covering nearby places at location $p_t$ which gets constantly updated using Equation 1, external memory integrates local maps in long episodes and *only* updates the local spatial representations centered at $p_t$. This is reasonable because the agent frequently pays attention to its nearby environments ($m_t$) but receive less information from places far away due to its limited sensing capabilities, for example, limited field of view of onboard cameras.

Different from Neural Turing Machine [8] where the memory slots are arranged sequentially, our addressable memory, of size $F \times H \times W$, is indexed by spatial coordinates $\{(i,j) : i \in \{1, 2, ..., H\}, j \in \{1, 2, ..., W\}\}$ with memory vector $M(i,j)$ of size $F$ at location $(i,j)$. Because ESM is often expressed in the coordinate frame of the agent itself, we use location-based addressing mechanism and the locations of reading or writing heads are fixed to be always in the center of the memory. At time step $t$, same as BVU, the external memory in the GU module is updated by transforming all the past spatial information based on the egomotion $A_{\theta,d,l,t}$.

We formulate the returned reading vector $a_{h,w}$ as

$$a_{h,w} = \left\{ M(i,j) : i \in \left\{\frac{H}{2} - \frac{h}{2}, ..., \frac{H}{2} + \frac{h}{2}\right\}, \\ j \in \left\{\frac{W}{2} - \frac{w}{2}, ..., \frac{W}{2} + \frac{w}{2}\right\} \right\} \quad (4)$$

where the memory patch covers the area of memory vectors with width $w$ and height $h$. We simplify the writing mechanism for GU and use Equation 4 for the writing vector $b_{h,w}$. The external memory can then be written as:

$$M_t = (1 - R) * S(M_{t-1}, A_{\theta,d,l,t}) + R * b_{h,w}, \quad (5)$$

where $R$ is a pre-defined binary attention mask and value 1 on the mask denotes local area $m_t$ of fixed size $w$ and $h$.

In our case, two writing heads and two reading heads are necessary to achieve the following: (1) one reading head returns the memory vectors $m_{t-1}$ in order for BVU to predict $m_t$ using Equation 1; (2) GU performs updates by writing the predicted local accumulative space representation $m_t$ back into the memory to construct the global map in the egocentric coordinate; (3) GU keeps track of the visited places by writing the discriminative representation $h(I_t)$ at the center of the egocentric global map denoted as $(\frac{H}{2}, \frac{W}{2})$; (4) GU returns the memory vectors near to the current location for loop closure classification where the size of the searching area is parameterized by $w$ and $h$. We simplify the interaction between local space representations $m_t$ and $m_{t-1}$ with GU and set $w$ and $h$ to be the same size as $m_t$ and $m_{t-1}$.

## F. Training and Implementation Details

We train ESMN by stochastic gradient descent with learning rate $0.002$ and momentum $0.5$. Adam Optimizer [17] is used. At each time step, the ground truths are provided: local map, egomotion and loop closure classification label. We first train each module separately using L2 distance loss between predicted local maps and the ground truth in BVU and triplet loss in PU, and then load these pre-trained networks into ESMN. We then fine-tune the pre-trained ESMN on 2D3DS dataset [1] for only 1 epoch. The input frame size is $3 \times 64 \times 64$. We normalize all the input RGB images to be within the range $[-1, 1]$. The discriminative representation $h(I_t)$ is of dimension 128. The size of the local space representation $m_t$ is $h \times w = 32 \times 32$ covering the area $7.68 meters \times 7.68 meters$ in the physical world whereas the global map (the addressable memory) is of size $H \times W = 500 \times 500$. We empirically set $\lambda = 0.5$ in BVU based on the performance in the validation set. The memory vector in GU is of size $F = 128$. We collect training data from simulation environments in Gazebo [18] and train ESMN in Torch [3]. Source codes for the simulation environments and our detailed ESMN architecture are available to download [1].

## IV. EXPERIMENTS

We design eight 3D mazes and evaluate ESMN on these virtual mazes in the robotic simulator, Gazebo [18]. We also conduct experiments on seven indoor areas in 2D3DS dataset [1] which is collected using Matterport Camera from real-world indoor buildings. Figure 3 shows example mazes and realistic indoor settings. They feature complex geometry and varieties of textures. Typically, these indoor environments carry rich semantic information and cover large physical areas in hundreds of $m^2$. In the loop closure classification example in Section IV-C, we create digits on walls as the unique features of specific pathways in mazes. We use the data in maze 1 to 5 for training and validation and the rest for testing. In 2D3DS dataset, we use standard training and testing splits as [1]: area 1, 2, 3, 4, 6 as training set and area 5 as test set. The ground truths for 2D top-view of environments are obtained by using 2D laser scanner attached to the agent. Depending on physical constraints of individual robots, egomotion limits in continuous action space could vary. In our case, we define the egomotion with rotation limit $[-10°, 10°]$ per time step (0.25 $s$) and translation limit $[-0.1, 0.1]$ in meters per time step (0.25 $s$).

---
[1] https://github.com/Mengmi/Egocentric-Spatial-Memory

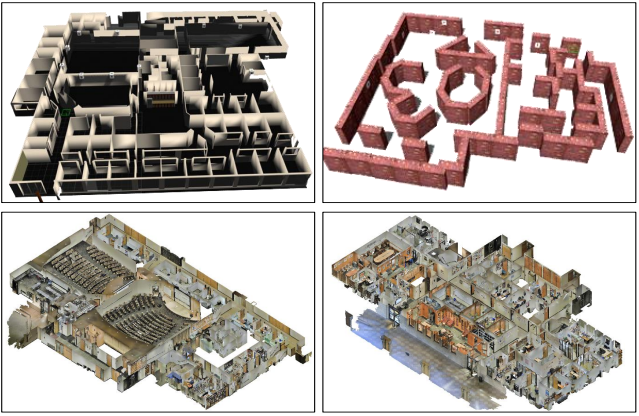

Fig. 3: Example simulation environments in 3D mazes and indoor areas. Mazes (Row 1) have different geometries and textures. Maze 1 is adopted from [6]. The realistic indoor areas from 2D3DS dataset [1] (Row 2) are collected using Matterport Camera from real-world indoor buildings.

### A. Boundary Vector Unit: Local Mapping

We express the proximity to physical obstacles by predicting 2D top-view free space representations and accumulate the local belief maps over time based on the pre-planned egomotions. Figure 4 shows example results of predicted accumulative free space representations over 32 time steps in indoor environments. Even in the complex office settings with tables and wall cabinets, ESMN could predict the free space (the narrow corridor) accurately.

We also provide quantitative analysis of our predicted local maps every 32 time steps in Figure 5 using Mean Squared Error (MSE), Correlation (Cor), Mutual Information (MI) which are standard image similarity metrics [24]. The predicted map and the ground truth map are normalized as gray scale images where each pixel value ranges [0,1] denoting free space as 1. Given the predicted local map $m_t$ and the ground truth map $g_t$, MSE can be defined as

$$MSE(m_t, g_t) = \frac{1}{hw} \sum_{i=1}^{w} \sum_{j=1}^{h} (m_t(i,j) - g_t(i,j))^2 \quad (6)$$

While MSE emphasizes the pixel-wise similarity, correlation and MI reflect the topological structure similarity. Specifically, correlation represents linear relationship between two distributions (maps) and MI measures how much information can be known from one distribution (map) given the other is known. Higher correlation and MI imply higher similarity of two distributions and hence, higher structural similarity between two maps. We choose 32 time steps for local map evaluations but this could be easily generalized to hundreds of time steps. At each time step, the predicted local maps are compared with the ground truth maps at $t = 32$. As the agent continues to explore in the environment, the area of the predicted free space expands leading to the decrease of MSE and the increase of correlation and MI in our test set. Moreover, we include two competitive baselines: (1) we train a binary classifier using 2D-CNN and classify whether each cell on grid maps is occupied or free (BiClassi) (2) we include a chance model by randomly assigning occupancy belief for each grid (chance). ESMN significantly outperforms all baselines, especially by around 0.02 in MSE, 0.1 in correlation and 0.15 in MI compared with BiClass. Compared with pixel-wise improvement in terms of MSE, our ESMN model makes more accurate inferences about topological structure of the world in terms of correlation and MI which validates that ESMN could accurately estimate the proximity to the physical obstacles relative to the agent itself and continuously accumulate the belief of free space representations and spatial connectivity.

### B. Place Unit: Loop Closure Detection

One of the critical aspects of ESM is the ability to recall and recognize a previously visited place independent of agent's orientation. We evaluate the learnt discriminative representations of visited places by comparing the current observation (anchor input) with all the past visited places in testing mazes. Figure 6a presents example pairs of observations when the loop closure is detected. Qualitative results imply that our ESMN can accurately detect loop closure when the agent is at the previously visited places irrespective of large differences between these camera views.

For quantitative evaluation, we formulate loop closure detection as a binary classification problem. Apart from the similarity comparison of observations in Equation 3, we consider two extra criteria for determining whether the place was visited: (1) the current position of the agent $p_t$ is near to the positions visited at the earlier times. We empirically set a threshold distance between the current and the recently visited locations based on the loop closure accuracy during training. We implemented it via a binary mask in the center of the egocentric global map where the binary states denote the accepted "closeness". (2) the agent only compares those positions which are far from the most recent visited positions to avoid trivial loop closure detection at consecutive time steps. It is implemented via another binary mask which tracks these recently visited places. These criterion largely reduce the false alarm rate and improves the searching speed during loop closure classification.

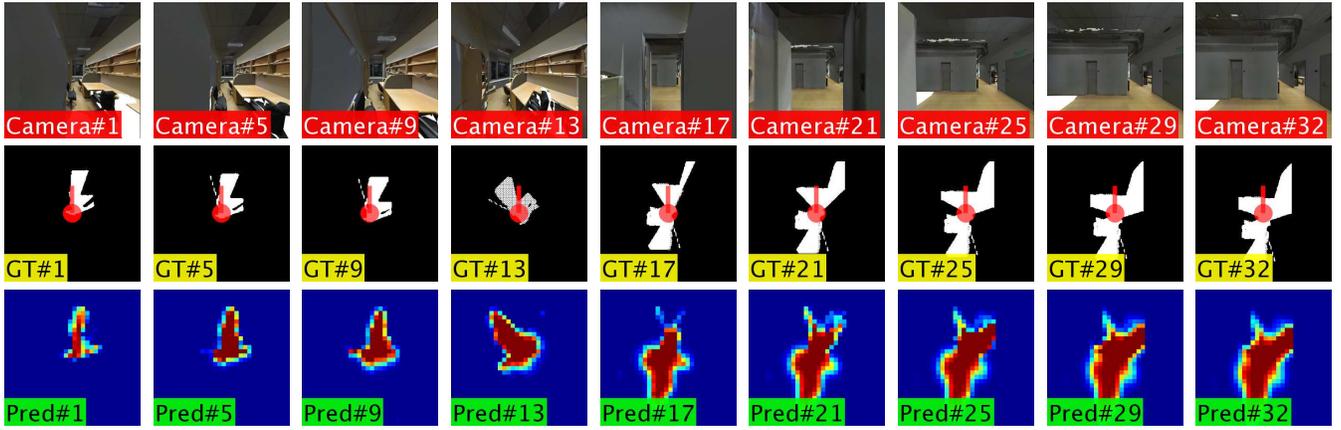

Fig. 4: Example results of predicted local maps over 32 time steps in indoor areas. Every 4 out of 32 frames are shown (left to right columns). Row 1 shows the camera views. Row 2 shows the ground truth with red arrows denoting the agent's position and orientation from the top view. The white region denotes free space while the black denotes unknown areas. Row 3 shows the corresponding top-view predicted local maps where the red color denotes higher belief of the free space.

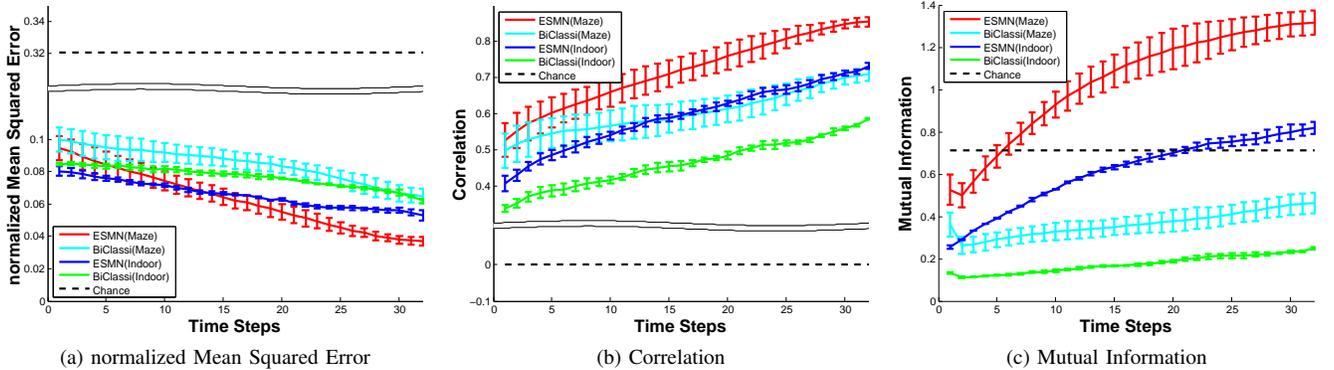

(a) normalized Mean Squared Error  (b) Correlation  (c) Mutual Information

Fig. 5: Evaluation of local mapper using normalized Mean Squared Error, Correlation, Mutual Information over 32 time steps in virtual mazes (red) and indoor areas (blue). The predicted local maps are compared with the ground truth at $t = 32$. We only show the evaluation results of local maps every 32 time steps but it could be easily generalized to hundreds of time steps. The models are: our model (ESMN) evaluated in mazes (red) and indoor areas (blue), binary classification baseline evaluated in mazes (green) and indoor areas (cyan), Chance (dash line). Smaller is better for MSE (1 is the worst). Larger is better for correlation and MI (0 is the worst for both). The error bars denote the standard deviation errors in the test sets.

In Figure 6c, we report the precision versus recall curve as we vary the classification threshold $\alpha$. Precision is defined as the ratio of true-positive loop-closure detections to the total number of detections. Recall is defined as the true-positive loop-closure detection rate. We also include the following baselines for comparison: (1) bag-of-words using SIFT descriptors as described by [33]; (2) pixel-wise comparison for pairs of camera views; (3) chance level (1/2 as binary classification). We observed that ESMN has the largest area under the curve (AUC) which demonstrates its robustness despite the variations of camera views.

### C. Grid Unit: Global Mapping

Although the focus of our paper is to propose an end-to-end deep network-based mapping framework given the ground truth egomotions in the continuous action space; in this section, we introduce measurement noise from motion sensors as an example and suggest a possible solution for our ESMN to adapt to noisy environments via loop closure classificaton and map correction.

If the loop closure classifies the current observation as "visited", GU eliminates the discrepancies on the global map by merging the two places together. The corrected map has to preserve the topological structure in the discovered areas and ensure the connectivity of the different parts on the global map is maintained. To realize this, we take three inputs in 3D convolution networks for map correction. The inputs are: (1) the local map predicted at the anchor; (2) the local map predicted at the recalled place; (3) all the past integrated global maps. To make the training targets, we perturb the sequence of pre-planned egomotions with noise and generate synthetic global maps with rotation and scaling augmentations. We minimize regression loss between the predicted maps and the ground truth. Figure 6b presents example results of the predicted global maps after map correction. It shows that the map gets corrected at $t = 448$ (Col1). Thus, the predicted global map (Row3) is structurally more accurate than the one without loop closure (Row2).

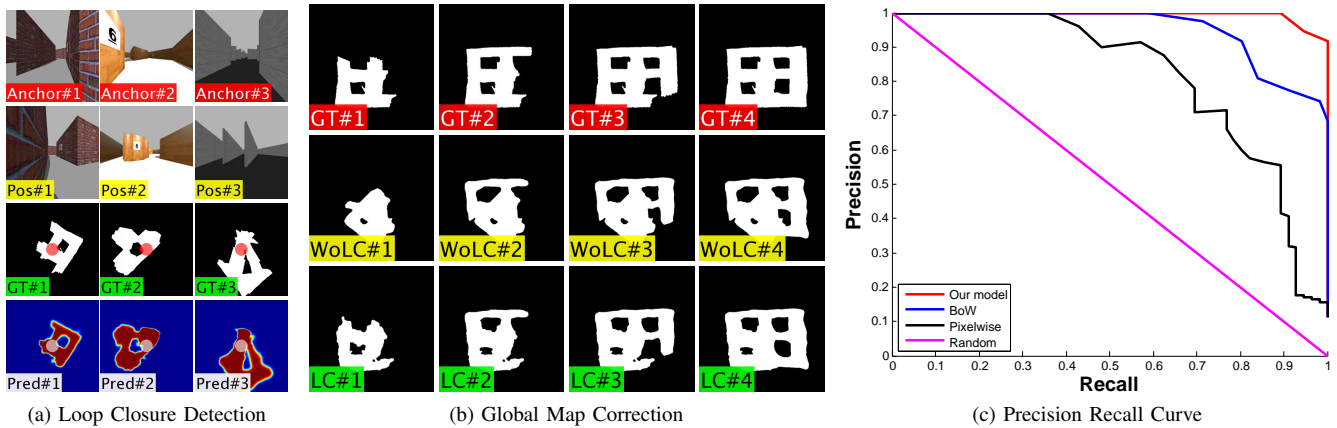

(a) Loop Closure Detection  (b) Global Map Correction  (c) Precision Recall Curve

Fig. 6: Place embedding in external memory in our proposed model after 15% noise is added in pre-planned egomotions in test mazes. **(a)** Example observation pairs when the loop closure is detected. Row 1 are the anchors (current camera views). Row 2 are the camera views from the previously visited places where the loop closure is detected. Row 3 show the agent's current locations (red circle) on the ground truth maps. Row 4 show the agent's locations (white circle) on the predicted map with ground truth poses. **(b)** Example results of constructed global maps in the world coordinate in Maze 6 across 1580 time steps. The topmost row shows the ground truth. Row 2 and Row 3 show the corresponding top-view accumulative belief of the predicted global maps without and with loop closure classification at $t = 448$ respectively. **(c)** Precision and recall curves of loop closure detection in test mazes. The models are: our model (red); bag-of-words with SIFT features (BoW, blue), pixel-wise comparison (Pixelwise, black) and chance (magenta). See Section IV-B for baseline descriptions.

|  |  | MSE | Cor | MI |
|---|---|---|---|---|
| Map Evaluation at t = 448 | | | | |
| Baselines | 3D-CNN | 0.09 | 0.50 | 0.15 |
|  | LSTM Direct | 0.09 | 0.48 | 0.19 |
| Ablated Models | HDU + BVU | 0.06 | 0.67 | 0.28 |
|  | HDU + BVU + PU + GU | **0.04** | **0.81** | **0.36** |
| Map Evaluation at t = 1580 | | | | |
| Baselines | 3D-CNN | t>model's capacity | | |
|  | LSTM Direct | 0.24 | 0.49 | 0.23 |
| Ablated Models | HDU+BVU | 0.06 | 0.83 | 0.58 |
|  | HDU + BVU + PU + GU | **0.04** | **0.91** | **0.72** |

TABLE I: Ablation study on the global map performance with 15% noise in pre-planned egomotions in Maze 6 using metrics in Section IV-A. It takes 1580 time steps to construct the global map in Maze 6. The loop closure detection and map correction happen at $t = 448$. From top to bottom, the models are: 3D-CNN baseline, LSTM baseline, our ablated model with PU and GU removed, our full model (ESMN). The best values are highlighted in bold.

We compare our global mapping capability with several baselines: (1) we take a sequence of camera views at all the previous time steps as inputs to predict the global map directly. We implement this by using a feed-forward 3D convolution neural network (3D-CNN). Practically, since it is hard to take all the past camera views across very long time period, we choose the input sequence with one representative frame every 15 time steps. (2) As ESM requires sustainable mapping over long durations, we create one more baseline by taking the sequence of camera views as inputs and using Long Short Term Memory architecture to predict global maps directly (LSTM Direct). To maintain the same model complexity, we attach the same 2D-CNN in our BVU module before LSTM and fully connected layers after LSTM. (3) To explore the effect of loop closure and thus map correction, we create one ablated model with PU and GU removed (HDU + BVU). (4) We present the results of our integrated architecture with loop closure classification and map correction enabled (HDU + BVU + PU + GU). We report the evaluation results in Table I using the metrics MSE, correlation and MI as introduced in Section IV-A.

We observe that ESMN surpasses all competitive baselines and ablated models. At $t = 448$, compared with 3D-CNN, there is decrease of 0.03 in MSE and increase of 0.17 in correlation and 0.09 in MI. The significant improvement verifies the important role of HDU. Moreover, the integration of local maps based on the egomotion makes the computation more flexible and efficient by feeding back the accumulative maps to the system for future time steps. In the second baseline (LSTM Direct), we observe that the performance drops significantly when it constructs global maps for longer durations. As GU serves as an external memory to integrate local maps, the baseline confirms GU has advantages over LSTMDirect in terms of long-lasting memory. To explore the effect of loop closure and thus map correction, we have the ablated model with PU and GU removed (HDU + BVU). Compared with our proposed architecture with all four modules enabled at $t = 448$ and $t = 1580$, the decreased performance validates that PU and GU are necessary to eliminate the errors during long-term mapping. In particular, compared with the improvement of 0.02 in MSE, we observe a more significant improvement of 0.08 in correlation and 0.14 in MI reflecting higher topological structure similarity. This has also been shown in Figure 6b where the global map is visually more structurally accurate after map correction though the averaged pixel-wise differences (MSE) between the global maps with and without loop closure classification are small.

## V. CONCLUSION

We propose an integrated deep neural network architecture for modeling egocentric spatial memory. Our learnt model demonstrates the capacity of constructing a top-down 2D spatial representation of the physical environments in the egocentric coordinate which could have many potential applications, such as path planning for robots. Our ESMN accumulates the belief about the free space by integrating egocentric camera views. To eliminate errors during mapping, ESMN also augments the local mapping module with an external spatial memory to keep track of the discriminative representations of the visited places for loop closure detection. We conduct exhaustive evaluation experiments in virtual mazes and realistic indoor environments. The experimental results demonstrate that our model could construct global maps accurately with the capability of detecting loop closure.

In the future, our model can be enhanced in several aspects: (1) our Grid Unit module requires fixed sizes for external memory; however, in practice, the global map with non-expandable memory size may be an issue for robots in large-scale exploratory missions. (2) In this paper, we restrict our discussions on occupancy grid map construction from egocentric camera views in a supervised approach; but one could attach our mapping framework with a down-stream path-planner to train a navigation agent via reinforcement learning (reward at goal location). (3) Our mapping framework can handle continuous action inputs. One application is to apply our mapping framework on a mobile robotic platform and make it interact with the control system in a real life navigation task.

## ACKNOWLEDGMENT

This work was supported by A*STAR JCO VIP grant 1335h00098 (REVIVE), National University of Singapore IDS R-263-000-C67-646, ECRA R-263-000-C87-133 and MOE Tier-II R-263-000-D17-112.